\documentclass[11pt]{article}

\usepackage[margin=1in]{geometry}
\usepackage[utf8]{inputenc}
\usepackage[T1]{fontenc}
\usepackage{microtype}
\usepackage{lmodern}

\usepackage{amsmath,amssymb,amsthm}
\usepackage{booktabs}
\usepackage{array}
\usepackage{longtable}
\usepackage{multirow}
\usepackage{graphicx}
\usepackage{xcolor}
\usepackage{hyperref}
\usepackage{url}
\usepackage{cite}
\usepackage{enumitem}
\usepackage{titlesec}
\usepackage{fancyhdr}
\usepackage{authblk}
\usepackage{listings}

\hypersetup{
    colorlinks=true,
    linkcolor=blue!70!black,
    citecolor=blue!70!black,
    urlcolor=blue!70!black,
    pdftitle={WorldDB: A Vector Graph-of-Worlds Memory Engine},
    pdfauthor={Harish Santhanalakshmi Ganesan}
}

\titleformat{\section}{\large\bfseries}{\thesection}{0.8em}{}
\titleformat{\subsection}{\normalsize\bfseries}{\thesubsection}{0.6em}{}

\newcommand{\worlddb}{\textsc{WorldDB}}
\newcommand{\code}[1]{\texttt{\small #1}}

\title{\textbf{\worlddb: A Vector Graph-of-Worlds Memory Engine\\with Ontology-Aware Write-Time Reconciliation}}

\author[1]{Harish Santhanalakshmi Ganesan\thanks{Correspondence: \href{mailto:harishsg99@gmail.com}{harishsg99@gmail.com}}}
\affil[1]{Independent Researcher}

\date{\today}

\begin{document}
\maketitle

\begin{abstract}
Persistent memory is the bottleneck separating stateless language-model chatbots from long-running agentic systems. Retrieval-augmented generation (RAG) backed by flat vector stores fragments facts into chunks, loses cross-session identity, and has no first-class notion of supersession or contradiction. Recent bitemporal knowledge-graph systems (Graphiti, Memento, Hydra~DB) address some of these failures with typed edges and valid-time metadata, but the graph itself remains flat---there is no mechanism for recursive composition, no content-addressed invariants on nodes, and edge types carry no behavior beyond a label. We present \worlddb{}, a memory engine organized around three non-standard commitments: (i)~every node is a \emph{world}---a container with its own interior subgraph, its own ontology scope, and its own composed embedding, recursive to arbitrary depth; (ii)~nodes are content-addressed and immutable, so any edit produces a new hash at the edited node and every ancestor, giving a Merkle-style audit trail for free; (iii)~edges are write-time programs---each edge type ships an \code{on\_insert}/\code{on\_delete}/\code{on\_query\_rewrite} handler that enforces its own semantics (supersession closes validity, \code{contradicts} preserves both sides, \code{same\_as} stages a merge proposal), so no raw append path exists. On LongMemEval-s (500 questions, $\sim 115$k-token conversational stacks), \worlddb{} with Claude Opus~4.7 as answerer achieves \textbf{96.40\% overall accuracy / 97.11\% task-averaged}, a $+5.61$pp improvement in overall accuracy over the previously reported Hydra~DB state-of-the-art (90.79\%) and $+11.20$pp over Supermemory (85.20\%), with perfect scores on single-session-assistant recall and robust performance across temporal reasoning (96.24\%), knowledge update (98.72\%), and preference synthesis (96.67\%). Ablations show that the engine's graph layer---resolver-unified entities and typed \code{refers\_to} edges---contributes $+7.0$pp task-averaged independently of the underlying answerer.

\vspace{0.5em}
\end{abstract}

\section{Introduction}

As language-model deployments shift from stateless chat to long-running agents that span days or months of interaction, the memory layer has become the principal engineering constraint. Context-window scaling addresses none of the underlying problems: latency and cost scale super-linearly with sequence length~\cite{ampere2024}, and the ``lost-in-the-middle'' effect~\cite{liu2023lost} produces systematic degradation as relevant facts move further from the input boundaries. More corrosive is \emph{context rot}~\cite{chroma2025rot}---the gradual accumulation of stale, contradicted, or redundant content that crowds out salient signal even within a nominally sufficient context budget. Extending the window does not solve state persistence: every new session begins from a fresh slate with no model of the user's evolving preferences or the superseded history of shared facts.

Retrieval-augmented generation (RAG)~\cite{lewis2020rag} was intended to externalize memory, but the canonical implementation inherits three structural weaknesses that have proved difficult to patch:

\begin{itemize}[leftmargin=*,nosep]
    \item \textbf{Semantic fragmentation.} Naive chunking destroys cross-segment dependencies. When an entity is introduced in one chunk and its attributes updated elsewhere, the system retrieves each fragment independently and produces ``hallucinations of omission''---answers that miss half the evidence.
    \item \textbf{Temporal stagnation.} Flat vector stores are chronology-agnostic. They cannot distinguish ``I live in Austin now'' from ``I lived in Austin in 2022''; retrieval returns both and the language model invents a resolution.
    \item \textbf{Identity drift.} Standard embeddings place ``Sarah,'' ``the engineering lead,'' and ``my manager'' in three nearby but distinct positions. Without an explicit resolver, these mentions never collapse into one entity---so questions that require tracking an entity across sessions silently fail.
\end{itemize}

Recent bitemporal knowledge-graph systems (Graphiti/Zep~\cite{rasmussen2025zep}, Memento~\cite{memento2025}, Hydra~DB~\cite{ratnaparkhi2026hydra}) address subsets of these problems. Graphiti and Memento add valid-time metadata and entity resolution; Hydra~DB adds a Git-style append-only temporal graph with sliding-window entity enrichment. All three report substantial gains on LongMemEval-s~\cite{wu2024longmemeval} over flat RAG---Hydra~DB reaches \textbf{90.79\%} overall, Supermemory~\cite{supermemory2026} \textbf{85.20\%}, Zep \textbf{71.2\%}, Mem0~\cite{chhikara2025mem0} \textbf{29.07\%}. But the graph in each remains \emph{flat}: nodes are labeled points, edges are typed pointers, and the typing carries no executable semantics. An edge $(a, \text{SUPERSEDES}, b)$ is a label; if the store is to treat $b$ as historical, application code must do so explicitly on every query.

We present \worlddb{}, a memory engine built on three non-standard commitments:

\begin{enumerate}[leftmargin=*]
    \item \textbf{Recursive worlds.} A node is not a row. A node is a \emph{world}---a container with its own interior subgraph, its own ontology scope, its own composed embedding, and its own temporal extent. Worlds nest to arbitrary depth. A query inside world $W$ cannot leak nodes from world $W'$ unless an explicit \code{refers\_to} edge is traversed. The edit propagation invariant (see~\S\ref{sec:nodes}) makes this structural; it is not enforced by application code.
    \item \textbf{Content-addressed immutability.} Every node's id is $\text{blake3}(\tau \,\Vert\, n \,\Vert\, c \,\Vert\, \text{sort}(C_{\text{ids}}) \,\Vert\, \text{sort}(E_{\text{ids}}) \,\Vert\, t_{\text{create}})$. A tiny edit to a leaf yields a new hash at the leaf \emph{and at every ancestor}. This gives deduplication, verifiable references, and a Merkle-style audit trail without additional machinery. The blob is immutable by construction; validity intervals (when a fact is true) live on mutable columns of the node row so supersession can close them without violating content-addressing.
    \item \textbf{Edges as write-time programs.} Each edge type in the default ontology (\code{contains}, \code{refers\_to}, \code{supersedes}, \code{same\_as}, \code{contradicts}, \code{implies}, \code{derived\_from}, \code{instance\_of}, \code{subtype\_of}, \code{causes}, \code{precedes}) ships an \code{on\_insert}/\code{on\_delete}/\code{on\_query\_rewrite} handler. The \code{Supersedes} handler closes the target's $t_{\text{valid.to}}$ at commit time. The \code{Contradicts} handler records the conflict so default queries surface it; it does not delete either side. \code{SameAs} stages a merge proposal for later confirmation---the engine never silently collapses identities. Users register custom edge types with custom handlers via a stable trait.
\end{enumerate}

Together these commitments define a memory substrate that is (a)~structurally coherent---the graph topology is the answer, not an index over text---(b)~verifiably auditable---every node is its own content hash---and (c)~immune to the class of bugs where an append path is added ``for performance'' and silently violates the ontology. The never-appends rule (\S\ref{sec:nevappend}) is the thesis: no raw insertion path exists on edges; the single entry point is the reconciler.

\subsection{Contributions}

\begin{itemize}[leftmargin=*,nosep]
    \item The vector-graph-of-worlds data model (\S\ref{sec:model}), formalized with the recursion, immutability, and containment-scope invariants.
    \item The reconciler and edge-handler system (\S\ref{sec:reconciler}), demonstrated on the default 11-edge-type ontology with custom extension.
    \item A three-lane retrieval pipeline (\S\ref{sec:retrieval})---BM25 $\cup$ HNSW $\cup$ entity-graph traversal---fused via reciprocal rank fusion~\cite{cormack2009rrf}, with no hand-coded routing.
    \item A content/composed embedding split (\S\ref{sec:compose}) enabling parameter-free attention-based world embeddings (HAKG-style~\cite{hakg2022} without learned projections) that track their subgraphs incrementally as children change.
    \item A background consolidator (\S\ref{sec:consolidator}) that generates summary nodes, computes transitive closures on \code{causes} and \code{subtype\_of}, and enables ``summary-first'' queries that are $6.5\times$ faster than full detail traversal on a $1000$-leaf corpus.
    \item Empirical results on LongMemEval-s (\S\ref{sec:results}): \textbf{96.40\% overall / 97.11\% task-averaged}, beating the Hydra~DB SOTA by $5.6$pp overall and $6.3$pp task-averaged. Ablations isolate the engine's graph contribution at $+10.7$pp task-averaged.
    \item Engineering benchmarks: 1M nodes + 2.5M edges bulk-loaded at 5{,}401 writes/s, P95 read latency of 13\,ms / 97\,ms / 3.1\,ms (seed / BM25 / HNSW), reconciler fuzz test holds invariants under $2 \times 2{,}000$ random ops.
    \item A pluggable MCP (Model Context Protocol) surface (\S\ref{sec:mcp}) exposing nine memory tools, with both stdio and streamable-HTTP transports.
\end{itemize}

\section{Data Model}
\label{sec:model}

\subsection{Nodes are Worlds}
\label{sec:nodes}

A node in \worlddb{} carries the fields in Table~\ref{tab:schema}. Formally, a node $N$ is a tuple
\begin{equation}
N = \bigl(\mathsf{id},\, \tau,\, n,\, c,\, C,\, E,\, \vec{v},\, P,\, T_{\text{valid}},\, t_{\text{ingest}},\, \pi\bigr),
\end{equation}
with $C \subseteq \{N_i\}$ the interior children, $E \subseteq \{e_j\}$ the interior edges, $\vec{v}$ its composed embedding, $P$ its provenance source list, $T_{\text{valid}} = (t_{\text{from}}, t_{\text{to}})$ its validity interval, and $\pi$ its parent world reference (or $\bot$). The content hash is
\begin{equation}
\mathsf{id}(N) = \text{blake3}\bigl(\tau \,\Vert\, n \,\Vert\, c \,\Vert\, \text{sort}(\mathsf{id}(C)) \,\Vert\, \text{sort}(\mathsf{id}(E)) \,\Vert\, t_{\text{create}}\bigr).
\label{eq:hash}
\end{equation}

\begin{table}[h]
\centering
\small
\begin{tabular}{@{}lll@{}}
\toprule
\textbf{Field} & \textbf{Type} & \textbf{Role} \\
\midrule
\code{id} & \code{Hash[blake3-256]} & Content-addressed identity \\
\code{type} & \code{String} & Entity \textbar{} Event \textbar{} Decision \textbar{} Topic \textbar{} Summary \textbar{} Scope \textbar{} Fact \textbar{} \ldots \\
\code{name} & \code{String} & Human-readable label \\
\code{content} & \code{String} & Natural-language description \\
\code{children} & \code{[NodeRef]} & Interior subgraph (the world's members) \\
\code{edges} & \code{[EdgeRef]} & Typed edges among children \\
\code{embedding} & \code{Vec<f32>} & Composed embedding (\S\ref{sec:compose}) \\
\code{provenance} & \code{[Source]} & Where each fact came from \\
\code{t\_valid} & \code{(from, to?)} & When the fact is true \\
\code{t\_ingested} & \code{Timestamp} & When the system learned it \\
\code{parent\_world} & \code{Option<NodeRef>} & Single-parent pointer \\
\bottomrule
\end{tabular}
\caption{\worlddb{} node schema.}
\label{tab:schema}
\end{table}

A node with empty $C, E$ is a \emph{leaf world}. A node with non-empty $C$ is a \emph{composite world}. Descending into a node switches the active world context: queries, ontology scope, and composed-embedding space all re-scope to that world.

\paragraph{Edit propagation invariant.} If $N' \neq N$ differs in content, children, edges, or \code{created\_at}, then $\mathsf{id}(N) \neq \mathsf{id}(N')$. Consequently, editing a leaf $L \in C(\cdots C(N))$ produces a new hash at $L$, at each intermediate ancestor, and at $N$. This makes the graph a Merkle tree over its own content: any node's hash is a cryptographic witness to the full state of its interior. Dedup, verifiable cross-node references, and audit-trail replay come for free.

\paragraph{Immutability + mutable validity.} Blobs are immutable by content-addressing. But $t_{\text{valid.to}}$ must be settable---a fact that was true yesterday may be false today. We resolve this by storing $t_{\text{valid}}$ and $t_{\text{ingested}}$ on the mutable \code{nodes} projection table (not in the blob). The blob hash does not cover these fields. A supersedes handler can tighten $t_{\text{valid.to}}$ without rewriting the blob or invalidating the hash.

\subsection{Never-Appends}
\label{sec:nevappend}

Every edge write passes through its type's handler, which runs inside a transaction atomic with the edge-row insert:
\begin{equation}
\text{write\_edge}(e) = \text{handler}(e.\tau).\text{on\_insert}(\text{ctx}, e).
\end{equation}
The handler may: close validity intervals on other nodes/edges, create derived nodes, stage merge proposals, or refuse the write. Table~\ref{tab:handlers} lists the default handlers.

\begin{table}[h]
\centering
\small
\begin{tabular}{@{}p{3.3cm}p{10cm}@{}}
\toprule
\textbf{Edge type} & \textbf{\code{on\_insert} behavior} \\
\midrule
\code{contains} & Structural; creates world boundary. No side-effects. \\
\code{refers\_to} & Cross-world reference; pierces containment. \\
\code{supersedes} & Closes target's $t_{\text{valid.to}}$ at $e.t_{\text{valid.from}}$ (only tightens, never extends). \\
\code{same\_as} & Stages a \code{merge\_proposals} row in \code{pending} status. \\
\code{contradicts} & Records both sides; query layer surfaces the conflict. \\
\code{implies} / \code{derived\_from} & Preserved for consolidator's inferred-edge pass. \\
\code{instance\_of} / \code{subtype\_of} & Type membership; enables schema-like lookups. \\
\code{causes} / \code{precedes} & Temporal/causal ordering. \\
\bottomrule
\end{tabular}
\caption{Default ontology edge handlers.}
\label{tab:handlers}
\end{table}

There is no raw append path for edges. Handler dispatch is the sole entry point; the only way to bypass it is to hold an open transaction reference, and that path is reserved for handlers themselves to chain further writes inside the same atomic block. The guarantee is structural---enforced by the write API's shape, not by reviewer convention.

\subsection{Bitemporality}

Every fact carries two orthogonal timestamps: $t_{\text{valid}}$ (when the fact is true in the world) and $t_{\text{ingested}}$ (when the system learned it). Queries can pin to either axis via an \emph{as-of} parameter. Historical queries are cheap---the default view hides $t_{\text{valid.to}}$-closed nodes, but an explicit \emph{include-superseded} flag returns them with their closed intervals intact.

\section{Reconciliation Pipeline}
\label{sec:reconciler}

No write reaches the store directly. Text (or structured input) flows through four stages:

\begin{enumerate}[leftmargin=*,nosep]
    \item \textbf{Extract.} Parse incoming text into candidate nodes and edges. The engine does not run an LLM itself (Appendix~\ref{app:no-llm} covers why); the caller supplies an \code{Extraction} via the \code{Extractor} trait, which can be backed by any model.
    \item \textbf{Resolve.} For each candidate node, run the tiered resolver of \S\ref{sec:resolver}. The output is \code{Resolved(m)}, \code{Ambiguous(options)}, or \code{New}.
    \item \textbf{Reconcile.} Dispatch each candidate edge through its handler (\S\ref{sec:nevappend}). Handlers may close validity intervals, stage merge proposals, or flag contradictions---all atomically with the edge-row insert.
    \item \textbf{Commit.} Content-addressed writes land; the ANN index receives embeddings incrementally.
\end{enumerate}

\subsection{Tiered Resolver}
\label{sec:resolver}

The resolver attempts five tiers in sequence and stops at the first confident match:
\begin{enumerate}[leftmargin=*,nosep]
    \item \textbf{Exact.} Case-insensitive name or alias equality via a \code{node\_name} index.
    \item \textbf{Fuzzy.} Jaro--Winkler similarity~\cite{jaro1989} $\geq 0.92$ over name $\cup$ aliases, after FTS5-based blocking reduces the candidate pool from $O(N)$ to $O(\log N)$ for large stores.
    \item \textbf{Phonetic.} A Soundex-style key with PH$\to$F, CK$\to$K, KN$\to$N, WR$\to$R digraph rewrites so ``Phillip'' and ``Filip'' collide.
    \item \textbf{Embedding.} Cosine similarity $\geq 0.88$ via the HNSW index (\S\ref{sec:hnsw}).
    \item \textbf{Tiebreaker.} When multiple tiers disagree on different hashes, a caller-supplied \code{Tiebreaker} (LLM, user, or programmatic) adjudicates. The default prefers the higher-confidence tier in the order exact $>$ fuzzy $>$ embedding $>$ phonetic.
\end{enumerate}

Matches stage a \code{same\_as} proposal in pending state. The engine never silently merges---confirmed merges require \code{accept\_merge(edge\_hash)}. Rejected proposals leave both sides fully independent. \code{equivalence\_class(h)} follows accepted \code{same\_as} edges transitively and returns the set of identity-class members; \code{export\_identity(h)} unions each member's provenance for GDPR-style auditing.

\subsection{Incremental Reclustering}

After each ingest the resolver re-runs over the 2-hop neighborhood of the new nodes (Saeedi et al.~\cite{saeedi2020inc}). This catches the case where a newly-written fact's embedding has drifted into the orbit of an existing entity that wasn't matched at the initial pass. Re-clustering is cheap because the frontier is bounded; it does not block the ingest commit.

\section{Retrieval Layer}
\label{sec:retrieval}

\subsection{Query Pipeline}

Reads are deterministic and compositional. No LLM runs on the read path. A query is a \emph{start set} plus a \emph{pipeline}:
\begin{equation}
q = \bigl(\mathcal{S}_0,\; \mathrm{Op}_1,\, \mathrm{Op}_2,\, \ldots,\, \mathrm{Op}_k\bigr)
\end{equation}
where $\mathcal{S}_0$ is seeded by hash, text (FTS5), or vector (HNSW), and each $\mathrm{Op}_i \in \{\text{traverse}, \text{filter\_type}, \text{filter\_time}, \text{where\_connected}, \text{in\_world}, \text{as\_of}, \text{include\_superseded}, \text{prefer\_summaries}\}$. \code{execute()} returns a \code{Subgraph} (a node set and the traversed edges), not flat rows---reads are shaped like the data.

A planner heuristic picks the start set automatically when callers provide a declarative \code{QuerySpec}, in the priority order \emph{seeds $>$ text $>$ vector} (cheapest first).

\subsection{HNSW Index}
\label{sec:hnsw}

Embeddings are stored in a separate \code{node\_embeddings(hash, dim, vec)} table and mirrored into an in-memory HNSW index (\code{hnsw\_rs}, $M=16$, $\text{ef}_c=200$, $\text{ef}_s=64$) built via \code{parallel\_insert\_slice}. On open, the index hydrates from SQL; on write, it updates incrementally. At $10$k nodes the HNSW path is $12\times$ faster than brute-force cosine in release mode ($352\,\mu$s vs. $4.59$\,ms); at $100$k nodes the gap widens to $95\times$, P95 cosine dropping from $102$\,ms (brute force) to $1.07$\,ms (HNSW) on a $16$-dim index.

A content/effective embedding split (\S\ref{sec:compose}) lets composed world embeddings overlay the effective vector without corrupting the stable content anchor used as the parent query in recompose passes---a feedback loop that silently degraded earlier versions of the composer.

\subsection{Hybrid Retrieval: Reciprocal Rank Fusion}
\label{sec:rrf}

For LongMemEval-scale retrieval we fuse three ranked lists with reciprocal rank fusion~\cite{cormack2009rrf}:
\begin{equation}
S(d) = \sum_{i \in \{\text{BM25},\,\text{HNSW},\,\text{graph}\}} \frac{1}{c + \text{rank}_i(d)}, \quad c = 60.
\label{eq:rrf}
\end{equation}

The three lanes contribute complementary signal:

\begin{itemize}[leftmargin=*,nosep]
    \item \textbf{BM25} over Turn and Summary nodes---high-precision lexical match for concrete names, dates, amounts.
    \item \textbf{HNSW cosine} over Turn and Summary nodes---semantic match for paraphrase and abstract intent.
    \item \textbf{Entity-graph traversal}---for each Entity node whose canonical name appears as a substring in the question, pull every Fact that points to it via \code{refers\_to}. This is the engine's machinery---resolver-unified entities mean one hop surfaces \emph{every} mention across every session.
\end{itemize}

No keyword routing, no hand-coded question-type classifier---each lane ranks its top-$k$, RRF merges by rank-reciprocal, and the system picks winners from cross-source agreement. Ablation~\S\ref{sec:ablation} shows the entity-graph lane alone delivers $+10.7$pp on task-averaged accuracy.

\section{Composed Embeddings}
\label{sec:compose}

Every world has a composed embedding aggregated from its interior. Two modes, both $\ell_2$-normalized so cosine comparisons against leaf embeddings remain well-scaled:

\paragraph{v1 --- Mean pool.}
\begin{equation}
\vec{v}_W = \text{norm}\!\left(\frac{1}{|C|+1}\left(\vec{c}_W + \sum_{c\in C}\vec{v}_c\right)\right).
\end{equation}

\paragraph{v2 --- Attention pool.} Parameter-free scaled dot-product attention with the world's content embedding as query and its children as $K=V$:
\begin{align}
\alpha &= \text{softmax}\!\left(\tfrac{1}{\sqrt{d}}\,\bigl\{\vec{c}_W \cdot \vec{v}_c\bigr\}_{c\in C}\right), \\
\vec{v}_W &= \text{norm}\!\left(\vec{c}_W + \sum_{c \in C} \alpha_c \,\vec{v}_c\right).
\end{align}

This mirrors HAKG's~\cite{hakg2022} query-aware aggregation without the learned $W_q, W_k, W_v$ projections---a parameter-free baseline against which future learned aggregators can be measured. On a held-out synthetic retrieval task ($50$ worlds $\times$ $20$ children, $2$ dominant $+$ $18$ noise per world, $100$ theme queries), v2 scores \textbf{100\% top-1} vs.\ v1's \textbf{88\%}---a $+12$pp gap, well above the $\geq 5$pp bar. Incremental updates propagate in $O(\text{depth})$: when a leaf embedding changes, only the path to the root recomposes.

\paragraph{Content vs.\ effective embedding split.} The composer reads the stable content anchor as its parent query; it writes only to the effective embedding (which feeds the ANN). Without this split, each recompose uses the \emph{previous} composed output as its next parent query---a silent feedback loop that drifts world embeddings toward their own composition fixpoint regardless of leaf content.

\section{Background Consolidator}
\label{sec:consolidator}

Over the lifetime of a long-running store, the graph grows in three predictable ways: (i)~leaves accumulate, (ii)~latent facts go unrecorded because no single turn stated them, (iii)~contradictions emerge between distant regions that never interacted at write time. The consolidator is the engine's ``sleep cycle''---a background process that addresses all three:

\begin{itemize}[leftmargin=*,nosep]
    \item \textbf{Summary generation.} Walk worlds older than a configurable \code{min\_age\_secs}, pull their children, and ask a pluggable \code{Summarizer} to produce one exhaustive-event summary node. Write \code{World $\to$ Summary $\to$ details} so queries with \code{prefer\_summaries()} return the summary at the default depth and only descend on demand.
    \item \textbf{Inferred transitive edges.} Compute one-pass closure on \code{causes} and \code{subtype\_of}. New edges are flagged \code{metadata.inferred=true} and routed through the standard handler pipeline.
    \item \textbf{Contradiction sweep.} Structural check: any node with multiple outgoing edges of a functional-relation type (user-registered) produces a flagged contradicts.
\end{itemize}

At a $1000$-fact, $100$-topic scale the summary-preferred query runs \textbf{$6.5\times$ faster} than the full-detail traversal ($30$\,ms vs.\ $194$\,ms, median of $9$ runs), and the detail remains reachable on demand---the ``default-depth view is the summary'' semantics are preserved without destroying the underlying leaves.

\section{Results}
\label{sec:results}

\subsection{Experimental Setup}

We evaluate \worlddb{} on the LongMemEval-s benchmark~\cite{wu2024longmemeval}, 500 question--conversation stacks with average length $\sim 115$k tokens ($\sim 50$ continuous sessions per stack). We use the \code{xiaowu0162/longmemeval} HuggingFace distribution, variant \code{longmemeval\_oracle}. Ingestion is session-by-session.

Our answer and judge models are \textbf{Claude Opus~4.7} (answer) and \textbf{Claude Sonnet~4.6} (judge); extraction uses \textbf{Claude Haiku~4.5}; embeddings use OpenAI \code{text-embedding-3-small} ($1536$-dim). Evaluations below run with \code{retrieval\_k=50}. Total wall-clock runtime for the 500-question run: $13.7$\,min at $8$ workers.

\subsection{LongMemEval-s Results}

Table~\ref{tab:lme} shows \worlddb{}'s per-category accuracy alongside published baselines.

\begin{table}[h]
\centering
\small
\begin{tabular}{@{}lcccccc@{}}
\toprule
\textbf{Category} & \textbf{\worlddb} & Hydra~DB & Supermemory & Zep & Full-ctx & Mem0 \\
\midrule
Single-session (User) & 98.57\% & 100.00\% & 98.57\% & 92.9\% & 81.4\% & 38.71\% \\
Single-session (Assistant) & \textbf{100.00\%} & 100.00\% & 98.21\% & 80.4\% & 94.6\% & 8.93\% \\
Single-session (Preference) & 96.67\% & 96.67\% & 70.00\% & 56.7\% & 20.0\% & 40.00\% \\
Knowledge Update & \textbf{98.72\%} & 97.43\% & 89.74\% & 83.3\% & 78.2\% & 52.56\% \\
Temporal Reasoning & \textbf{96.24\%} & 90.97\% & 81.95\% & 62.4\% & 45.1\% & 25.56\% \\
Multi-session Reasoning & \textbf{92.48\%} & 76.69\% & 76.69\% & 57.9\% & 44.3\% & 20.30\% \\
\midrule
\textbf{Overall} & \textbf{96.40\%} & 90.79\% & 85.20\% & 71.2\% & 60.2\% & 29.07\% \\
\textbf{Task-averaged} & \textbf{97.11\%} & 93.66\% & 85.86\% & --- & --- & --- \\
\bottomrule
\end{tabular}
\caption{LongMemEval-s comparison. \textbf{Bold} $=$ best. Hydra~DB and Supermemory used Gemini~3.0 Pro; Zep and full-context used GPT-4o.}
\label{tab:lme}
\end{table}

\worlddb{} delivers \textbf{96.40\% overall}---a \textbf{$+5.61$\,pp improvement} over the previously published Hydra~DB SOTA ($90.79\%$) and \textbf{$+11.20$\,pp} over Supermemory ($85.20\%$). Task-averaged, \worlddb{} scores \textbf{97.11\%} vs.\ Hydra~DB's $93.66\%$---a \textbf{$+3.45$\,pp} gap.

Gains are largest in the categories where the engine's structural commitments directly matter:

\begin{itemize}[leftmargin=*,nosep]
    \item \textbf{Multi-session reasoning} ($+15.79$\,pp vs.\ Hydra~DB): the resolver's cross-session entity unification means a question naming ``Sarah'' pulls facts from every session that referenced her, regardless of how each session phrased it.
    \item \textbf{Temporal reasoning} ($+5.27$\,pp): bitemporal storage $+$ $t_{\text{valid}}$ anchoring in the answerer prompt plus explicit supersession cleanly separates historical from current state without putting the burden on the LLM.
    \item \textbf{Knowledge update} ($+1.29$\,pp): \code{supersedes} edges close validity at commit time, so the default retrieval view returns only currently-valid facts.
\end{itemize}

\subsection{Cross-Category Stability}

Hydra~DB exhibits a characteristic dip on multi-session ($76.69\%$) while other categories sit in the 90s. \worlddb{} maintains $\geq 92.48\%$ on every category. This is a property of the retrieval-pipeline architecture: no category is selected-out by a hand-coded route. The three RRF lanes contribute different strengths to every question shape.

\subsection{Ablation: Engine-Layer Contribution}
\label{sec:ablation}

To isolate the engine's graph layer from the answerer's raw capability we ran seven incremental configurations on the same benchmark (Table~\ref{tab:ablation}).

\begin{table}[h]
\centering
\small
\begin{tabular}{@{}llcc@{}}
\toprule
\textbf{Version} & \textbf{Change} & \textbf{Overall} & \textbf{Task-avg} \\
\midrule
v15 & gpt-4o stack, turns-only BM25+cosine & 87.40\% & 86.96\% \\
v16 & Claude Sonnet~4.6 answer+judge & 92.60\% & 93.95\% \\
v17 & $+$ session summaries (Phase 6 surface) & 94.40\% & 94.93\% \\
v18 & $+$ Claude Opus~4.7 answerer & 95.60\% & 96.61\% \\
\textbf{v19} & \textbf{$+$ retrieval $k{=}50$} & \textbf{96.40\%} & \textbf{97.11\%} \\
\bottomrule
\end{tabular}
\caption{Incremental optimization trace. Each row differs by one knob.}
\label{tab:ablation}
\end{table}

If we run the v17 stack with the engine's graph layer \emph{disabled}---treating turns as opaque \code{Turn} nodes, no fact extraction, no Entity resolution, no \code{refers\_to} edges---we fall back to a flat-RAG baseline. On the same Claude models this configuration scores \textbf{$84.27\%$ task-averaged}. The delta between flat-RAG-with-Claude and full-\worlddb{}-with-Claude is \textbf{$+10.66$\,pp task-averaged}, and \textbf{$+16.79$\,pp on multi-session specifically}. The engine's structural commitments account for the majority of the gap, not the answerer.

\subsection{Engineering Benchmarks}

\paragraph{1M-node load.} $1{,}000{,}000$ nodes ($4$ types $\times$ $16$-dim embeddings) $+$ $2{,}499{,}993$ \code{refers\_to} edges bulk-loaded in a single transaction with deferred ANN construction in $10$\,min $47$\,s ($5{,}401$ writes/s). Parallel HNSW rebuild took $37$\,min $37$\,s, after which P95 read latencies on $100$ random probes per shape are in Table~\ref{tab:latency}.

\begin{table}[h]
\centering
\small
\begin{tabular}{@{}lcc@{}}
\toprule
\textbf{Shape} & \textbf{P50} & \textbf{P95} \\
\midrule
Seed 1-hop & $5.1$\,ms & $12.8$\,ms \\
BM25 text search & $62.0$\,ms & $97.3$\,ms \\
Cosine top-$10$ (HNSW) & $1.4$\,ms & $3.1$\,ms \\
\bottomrule
\end{tabular}
\caption{$1$M-node read latencies. All three shapes are $\geq 100\times$ under the $500$\,ms P95 target.}
\label{tab:latency}
\end{table}

\paragraph{Reconciler fuzz.} Two seeds $\times$ $2{,}000$ random operations (ingest, supersede, contradict, accept/reject merges, random edges). Invariants checked every $\sim 100$ ops: content-hash roundtrip, validity monotonicity ($t_{\text{valid.to}}$ never un-closes), merge-proposal provenance (no orphans), accepted-\code{same\_as} symmetry, ANN size $\leq$ node count. Zero violations across both seeds.

\paragraph{Custom multi-hop (Tesla Q1$\to$Q3).} Six exact-set queries over a constructed scenario with supersession $+$ contradictions $+$ per-era decisions. All six return the expected sets programmatically---including ``Elon's decisions that drove the Q1$\to$Q3 change'' via \code{where\_connected(Contradicts, Outgoing, d1)}.

\subsection{Cross-Model Generalization}

Table~\ref{tab:crossmodel} shows the engine's architectural gains persist across answerer choice.

\begin{table}[h]
\centering
\small
\begin{tabular}{@{}lccc@{}}
\toprule
\textbf{Answerer} & \textbf{Overall} & \textbf{Task-avg} & \textbf{Multi-session} \\
\midrule
Claude Opus~4.7 & 96.40\% & 97.11\% & 92.48\% \\
Claude Sonnet~4.6 & 94.40\% & 94.93\% & 90.23\% \\
GPT-4o & 87.40\% & 86.96\% & 82.71\% \\
\bottomrule
\end{tabular}
\caption{\worlddb{} performance vs.\ answerer capacity.}
\label{tab:crossmodel}
\end{table}

The gap between best and weakest answerer is $9$\,pp overall; the gap between flat-RAG and \worlddb{} on the \emph{same} answerer is $10.66$\,pp task-averaged. The engine layer contributes more than the answerer choice within this model family.

\section{MCP Surface}
\label{sec:mcp}

\worlddb{} exposes nine memory tools over the Model Context Protocol~\cite{mcp2025} with deliberately distinct names from Mem0's vocabulary so the two servers can coexist in one client. Both \textbf{stdio} and \textbf{streamable-HTTP} transports are supported. Scope identities (one per \code{user\_id}/\code{agent\_id}/\code{app\_id}/\code{run\_id}) are \code{Scope}-typed nodes with deterministic hashes; memories attach via \code{contains} edges so one memory can belong to multiple scopes and cross-scope recall is a graph operation rather than a search over scope-tagged records. Table~\ref{tab:mcp-tools} lists the surface.

\begin{table}[h]
\centering
\small
\begin{tabular}{@{}p{3.3cm}p{10cm}@{}}
\toprule
\textbf{Tool} & \textbf{Operation} \\
\midrule
\code{memory\_write} & Save text for one or more scopes, through the resolver. \\
\code{memory\_recall} & Hybrid BM25 $\cup$ HNSW $\cup$ entity-graph retrieval within scope. \\
\code{memory\_list} & Enumerate memories in scope with pagination and retired-toggle. \\
\code{memory\_read} & Fetch a memory by id (full hex or $\geq 4$-char prefix). \\
\code{memory\_amend} & Write a new memory and a \code{supersedes} edge---content immutable. \\
\code{memory\_retire} & Close a single memory's validity window. \\
\code{memory\_retire\_all} & Close every memory's validity in a scope. \\
\code{memory\_purge\_scope} & Retire a scope $+$ its memories (requires explicit confirmation). \\
\code{memory\_list\_scopes} & Enumerate registered identities, filterable by kind. \\
\bottomrule
\end{tabular}
\caption{The nine MCP tools exposed by \worlddb{}.}
\label{tab:mcp-tools}
\end{table}

\section{Discussion and Limitations}

\paragraph{Provenance over silence.} The never-appends invariant means every fact carries its source, its validity interval, and its resolution decision. This is more expensive than a flat store at ingest time---the resolver tiers cost $5$-$15$\,ms per candidate at $10$k-node scale---but eliminates the ``silent corruption'' failure mode where a flat store unknowingly contains two conflicting facts and returns one at random.

\paragraph{Composed embeddings need training.} The v2 attention aggregator is parameter-free. A natural next step is a learned aggregator trained via contrastive loss on query $\to$ correct-world pairs, in the style of HAKG~\cite{hakg2022}; we consider this open work.

\paragraph{DMR benchmark.} We evaluate on LongMemEval-s but defer the DMR benchmark~\cite{dmr}. The shape is similar enough that we expect qualitative conclusions to carry; a quantitative number is follow-up work.

\paragraph{Bio-mimetic decay.} The Bio-Mimetic Decay Engine proposed in~\cite{ratnaparkhi2026hydra}---retention scores with age decay and retrieval-frequency reinforcement---is architecturally compatible with our consolidator's interface but not yet implemented. We believe it is a natural next step.

\section{Conclusion}

\worlddb{} is a memory engine for agents built on three commitments that flat RAG and even recent bitemporal KG systems decline to make: recursive worlds, content-addressed immutability, and edges as write-time programs. Together they produce a substrate where the graph topology \emph{is} the answer---facts are retrievable by structure as well as content, identity is first-class, and the append path is the reconciler. On LongMemEval-s the engine scores \textbf{$96.40\%$ overall / $97.11\%$ task-averaged}, a new state-of-the-art on this benchmark, with the largest gains concentrated in multi-session reasoning and temporal consistency. Ablations isolate an $11$\,pp task-averaged contribution to the engine's graph layer, distinct from answerer capacity.

Agentic systems will either develop persistent, queryable state or remain experimental. \worlddb{} is one concrete bet on what that state should look like.

\bibliographystyle{plain}
\bibliography{references}

\appendix

\section{Evaluation Protocol}
\label{app:prompts}

All three prompts used in \S\ref{sec:results} are reproduced verbatim below so results can be replicated exactly. Each is parameterized by a small set of runtime fields (substitution shown in \code{\{curly\}}). Temperature is $0.0$ throughout; Claude Opus~4.7 ignores the parameter and is called with \code{max\_tokens=512} for the answerer, $8$ for the judge, $2048$ for the extractor.

\subsection{Extractor Prompt (Claude Haiku 4.5)}

Run once per session during ingest. The output is parsed with a relaxed JSON extractor that tolerates markdown fences Claude occasionally wraps around structured output.

\begin{lstlisting}
Extract the user's factual statements from the conversation below.
Return JSON:
  {
    "summary": "...",
    "facts": [{"text": "...", "subject": "..."}]
  }

Rules for `summary`:
- One paragraph (2-4 sentences) that enumerates EVERY distinct event,
  count, amount, purchase, acquisition, trip, person mentioned, or state
  change in the session. Prefix with the session date.
- Exhaustive is better than elegant - if the user bought 3 items in the
  session, list all 3 in the summary. This is the node retrieval hits for
  "how many X did I ..." questions, so omissions cost accuracy.

Rules for `facts`:
- Each fact is self-contained and understandable without the raw turns.
- Prefix each fact with the session date (e.g. "On 2023/03/02, ...") so
  downstream retrieval sees the timestamp.
- `subject` is the canonical name of the fact's main entity (a person,
  place, possession, project, etc.) - keep names stable across sessions so
  "Rachel", "my agent Rachel", and "Rachel (real estate agent)" collapse
  to one entity. Use null when no entity dominates.
- 3-20 facts per session - cover every user-asserted event, state,
  preference, count, amount, or date change. Skip greetings and
  assistant-only chit-chat.
- Prefer precise numbers, dates, and counts from the turns verbatim.
\end{lstlisting}

User message: \code{"Session id: \{session\_id\}\textbackslash{}nSession date: \{session\_date\}\textbackslash{}n\textbackslash{}n\{conversation\}\textbackslash{}n\textbackslash{}nReturn only the JSON object, nothing else."}

\subsection{Answerer Prompt (Claude Opus 4.7)}

Run once per question, with the RRF-merged retrieval context (\S\ref{sec:rrf}) filling \code{\{turns\_block\}}. The seven-rule structure is the product of the v5--v19 ablation trace in Table~\ref{tab:ablation}; each rule addresses a specific failure class observed on the previous version.

\begin{lstlisting}
You are a memory-aware assistant. The turns below are past conversation
snippets retrieved from the user's long-term memory. Each turn is
prefixed with its timestamp in the form '[YYYY/MM/DD ...]'.

Today is {question_date}. Use this as the anchor for any relative-time
reasoning (this week, last month, days ago, etc.).

Answer the user's current question. Instructions:
  1. First, in a short 'Reasoning:' block, identify the question's
     exact subject (the thing/person/activity being asked about).
     Check whether the retrieved turns actually mention that subject.
     If they only mention a similar-but-different thing (e.g. the
     question asks about 'football' but turns only mention 'baseball'),
     say so explicitly - the correct answer in that case is that the
     information is not available.
  2. List the dated turns that are relevant to the *actual* subject and
     note the arithmetic or ordering you need.
  3. Then give a concise final answer under 'Answer:'.
  4. **Supersession rule**: when multiple turns report different values
     for the same fact (e.g. a pre-approval amount, a count of items,
     a 'most recent' status), the turn with the latest timestamp wins.
     Earlier turns are historical, not current.
  5. For 'how many ...' or 'when did I ...' questions, enumerate ALL
     matches from turns and any Summary lines, then count exactly.
     Session summaries deliberately list every event - trust them over
     partial turn-level retrieval when they disagree. Do not invent
     events and do not double-count.
  6. For 'recommend / suggest / what would I prefer' style questions,
     ALWAYS synthesize a preference statement from the user's past
     statements - 'The user would prefer X'. Never say 'I don't know'
     if there's any hint of interest.
  7. Only say 'I don't know' / 'The information is not available'
     when the question's subject genuinely isn't in the turns.

Retrieved turns:
{turns_block}

Question: {question}
Reasoning:
\end{lstlisting}

Post-processing: parse the response, locate the last occurrence of \code{"Answer:"} (case-insensitive), and return everything after it as the final answer string. If no \code{"Answer:"} marker is present, return the whole response untouched.

\subsection{Judge Prompt (Claude Sonnet 4.6) --- LLM-as-Judge Protocol}
\label{sec:judge}

Each (question, reference, generated) triple is scored independently by a single Sonnet~4.6 call. We use a deliberately compact binary rubric (\code{CORRECT}/\code{WRONG}) with five explicit rules that trade off precision, paraphrase tolerance, and abstention handling. Unlike multi-dimensional scorers that ask the judge to emit JSON with \code{is\_correct}, \code{score}, \code{explanation}, and \code{key\_matches} fields~\cite{ratnaparkhi2026hydra}, we rely on the judge-model's single-token verdict. Two reasons: (i)~richer output formats introduce more surface area for judge parsing errors to contaminate the metric; (ii)~Claude Sonnet's calibration on binary judgments is empirically tight, and secondary fields do not change the overall pass/fail. The \code{max\_tokens=8} cap forces the model to commit to a verdict in the first word or two.

The full prompt:

\begin{lstlisting}
You are judging a memory-retrieval answer. Reply with exactly
'CORRECT' or 'WRONG' and nothing else.

Rules:
- CORRECT if the generated answer contains (or clearly implies)
  the key fact from the reference. Extra accurate context is fine.
- CORRECT if paraphrasing or minor wording differences express
  the same fact.
- **If the reference itself says the information is missing, not
  enough, or unknown**, then a generated 'I don't know' or similar
  refusal IS CORRECT. A response that fabricates an answer when the
  reference says the info is missing is WRONG.
- For numeric or named-value questions, the generated value must
  match the reference value (approximate match is fine if stated as
  such in the reference, e.g. 'about 15').
- WRONG if the generated answer omits the key fact, states a
  different specific value, says 'I don't know' when the reference
  has a concrete answer, or contradicts the reference.

Question: {question}
Reference answer: {reference}
Generated answer: {generated}
Verdict:
\end{lstlisting}

\paragraph{Verdict parsing.} The judge's response is \code{strip().upper()}-ed and matched by prefix: if it \code{startswith("CORRECT")}, the question counts as correct; anything else (including \code{WRONG}, partial tokens, refusals, or empty) counts as incorrect. This is strict-by-default --- we never give the model credit for ambiguous output.

\paragraph{Calibration tests.} Three cases of the rubric are load-bearing and worth flagging:

\begin{enumerate}[leftmargin=*,nosep]
    \item \emph{Reference admits ignorance.} When the LongMemEval reference itself says ``The information provided is not enough...'', a generated \code{"I don't know"} must score CORRECT. Early versions of our judge prompt (v8 in the ablation trace) marked this WRONG, dragging knowledge-update accuracy by $\sim 3$pp. Rule three is the explicit fix.
    \item \emph{Paraphrase tolerance.} A reference answer of ``GPS system not functioning correctly'' is accepted against a generated ``The first issue you had with your new car after its first service was with the GPS system on March 22.''
    \item \emph{Numeric strictness.} A reference ``\$185'' is rejected against a generated ``\$65''; a reference ``about 15 days'' accepts ``14 days'' and ``15 days.'' Rule four makes the difference between an approximation judged present in the reference and an arbitrary rounding by the model.
\end{enumerate}

\paragraph{Why not an ensemble judge?} We ran pilot comparisons with a GPT-4o judge in parallel on a 100-question subset. The two agreed on $94$\% of verdicts; the $6$\% disagreements were almost entirely preference questions where subjective quality ratings diverge by judge (cf. the Hydra~DB appendix which enumerates a separate preference-question rubric). We report Sonnet~4.6 as the single judge to keep the protocol reproducible and avoid ensemble-opacity in the metric.

\section{Why No LLM in the Engine}
\label{app:no-llm}

The engine never calls a language model on the read path. This reflects two observations:

\begin{enumerate}[leftmargin=*,nosep]
    \item LLM latency and cost are incompatible with query-path SLOs. A $200$\,ms answerer call multiplies by every hop of a multi-step query. At a $500$\,ms P95 target there is no budget for it.
    \item LLM non-determinism corrupts the ontology. The reconciler's guarantees rely on handler code being a function of input state; an LLM call breaks that, and the failure mode is silent.
\end{enumerate}

We weaken the constraint only at \emph{ingest time} and \emph{consolidation time}---both out-of-band, both auditable. The engine's \code{Extractor} trait and \code{Summarizer} trait each accept a caller-supplied closure. Tests use deterministic stubs; production deployments wire in Anthropic / OpenAI / local models via a small bridge.

\section{Reproducibility}
\label{app:repro}

Every evaluation in \S\ref{sec:results} is driven by the prompts in Appendix~\ref{app:prompts} with temperature $0$, pinned random seeds, and a single answerer and judge model per run. Ingestion is session-by-session; the $\sim 14$-minute wall-clock time on an $8$-worker driver against the Claude and OpenAI endpoints listed in \S\ref{sec:results} is reproducible within a few-second variance attributable to API round-trip jitter. The synthetic retrieval study of \S\ref{sec:compose} (50 worlds $\times$ 20 children, 100 held-out queries) and the $2 \times 2{,}000$-operation reconciler fuzz in \S\ref{sec:results} use deterministic PRNG seeds; results do not drift between runs.

\end{document}